\documentclass[conference]{IEEEtran}
\IEEEoverridecommandlockouts
\usepackage{graphicx} 
\usepackage{algorithm}
\usepackage{amssymb}
\usepackage{url}
\usepackage{spconf,amsmath,epsfig}
\usepackage{amssymb} 
\usepackage{multirow} 
\usepackage{algpseudocode}
\usepackage[margin=1in]{geometry}
\usepackage{booktabs} 
\usepackage{tabularx} 
\usepackage{url}
\usepackage{siunitx} 
\usepackage{graphicx} 
\usepackage{algorithm}
\usepackage{multirow} 
\usepackage{amssymb} 
\usepackage[margin=1in]{geometry} 
\usepackage{algpseudocode}
\usepackage[breaklinks,colorlinks]{hyperref}
\usepackage{booktabs} 
\usepackage{tabularx} 

\usepackage{siunitx} 

\def\BibTeX{{\rm B\kern-.05em{\sc i\kern-.025em b}\kern-.08em
    T\kern-.1667em\lower.7ex\hbox{E}\kern-.125emX}}
\begin{document}\sloppy
\bstctlcite{IEEEexample:BSTcontrol}
\def\x{{\mathbf x}}
\def\L{{\cal L}}

\title{Explicit Correlation Learning for Generalizable Cross-Modal Deepfake Detection}

\name{Cai Yu$^{1,2,3}$, Shan Jia$^{3}$, Xiaomeng Fu$^{1,2}$, Jin Liu$^{1,2}$, Jiahe Tian$^{1,2}$, Jiao Dai$^{1\ast}$, Xi Wang$^{4}$, \\Siwei Lyu$^{3}$, Jizhong Han$^{1}$\thanks{$^{\ast}$Corresponding authors.\\Siwei Lyu is supported by National Science Foundation (NSF) project under grant SaTC-2153112.}}
\address{$^{1}$
\textit{Institute of Information Engineering, Chinese Academy of Sciences, Beijing, China}\\$^{2}$\textit{School of Cyber Security, University of Chinese Academy of Sciences, Beijing, China}\\$^{3}$\textit{University at Buffalo, State University of New York, NY, USA}\\$^{4}$\textit{Institute of Microelectronics, Chinese Academy of Sciences, Beijing, China}\\$\{$caiyu, fuxiaomeng, liujin, tianjiahe, daijiao, hanjizhong$\}$@iie.ac.cn,\\$\{$shanjia, siweilyu$\}$@buffalo.edu, $\{$wangxifine$\}$@gmail.com}

\maketitle

\begin{abstract}
With the rising prevalence of deepfakes, there is a growing interest in developing generalizable detection methods for various types of deepfakes. While effective in their specific modalities, traditional detection methods fall short in addressing the generalizability of detection across diverse cross-modal deepfakes. This paper aims to explicitly learn potential cross-modal correlation to enhance deepfake detection towards various generation scenarios. Our approach introduces a correlation distillation task, which models the inherent cross-modal correlation based on content information. This strategy helps to prevent the model from overfitting merely to audio-visual synchronization. Additionally, we present the Cross-Modal Deepfake Dataset (CMDFD), a comprehensive dataset with four generation methods to evaluate the detection of diverse cross-modal deepfakes. The experimental results on CMDFD and FakeAVCeleb datasets demonstrate the superior generalizability of our method over existing state-of-the-art methods. Our code and data can be found at \url{https://github.com/ljj898/CMDFD-Dataset-and-Deepfake-Detection}.
\end{abstract}
\section{Introduction}
\label{sec:intro}

Amidst the rapid advancement of AI-generated content (AIGC) technologies, the landscape of deepfakes has evolved to encompass a broader spectrum of forms. Previous deepfake technologies predominantly concentrated on manipulating individual expressions or identities, such as face reenactment~\cite{nirkin2019fsgan,thies2016face2face} and face swap~\cite{korshunova2017fast,perov2020deepfacelab}. Yet, cutting-edge deepfake techniques like talking head generation~\cite{wang2021audio2head,zhou2020makelttalk} and lip-sync~\cite{prajwal2020lip,cheng2022videoretalking}, have expanded their capabilities to the realm of cross-modal synthesis. These sophisticated methods utilize audio inputs to generate corresponding talking videos of target individuals, seamlessly synchronizing the lip and head motions in the visual portrayal with the spoken content of the audio. These highly realistic deepfakes, capable of manipulating individuals into ``speaking" any specified content, pose a growing threat to authentic identity security, especially on social media, where their misuse potential is alarmingly pronounced.
 \begin{figure}[t]
 \centering
\includegraphics[width=1.0\columnwidth]{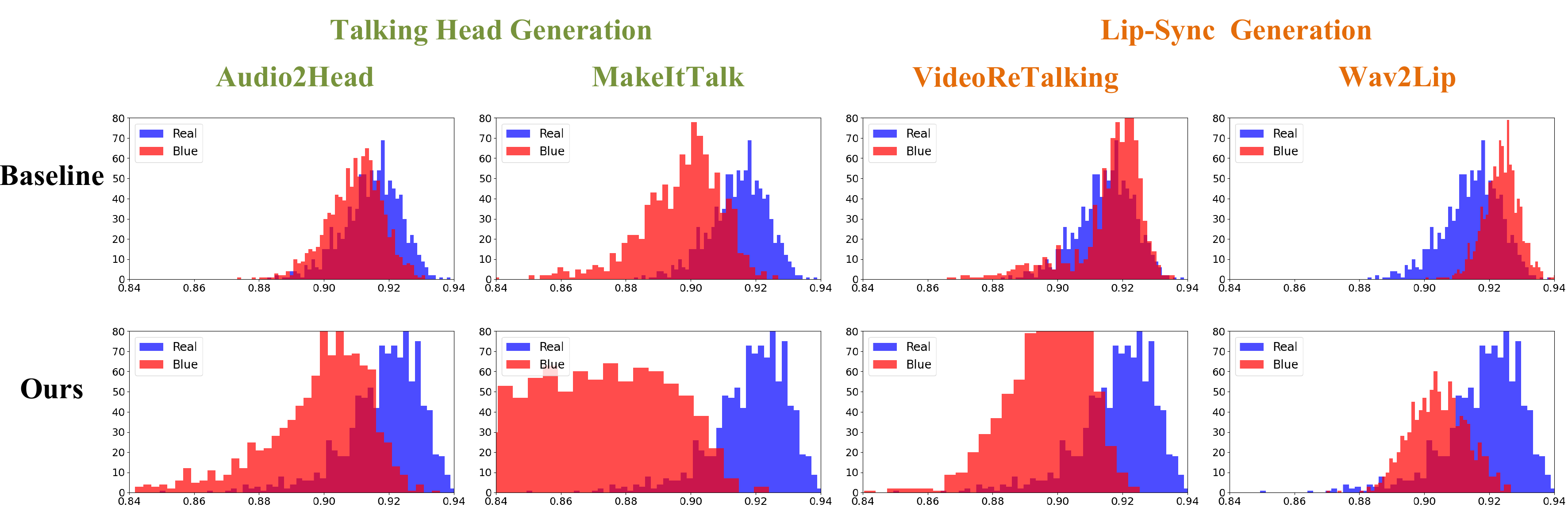}

\caption{Visualization of audio-visual correlation in various types of cross-modal deepfakes by calculating the cosine similarity between audio and visual embeddings (fake class in red and real in blue). The top row displays the correlation patterns captured by the baseline model~\cite{tao2021someone}, where the correlation exhibits varying intensities across different deepfake generation methods. Deepfakes generated by talking head methods tend to show relatively weaker audio-visual correlations. At the same time, lip-sync deepfakes, which are manipulated primarily in the lip region, demonstrate a higher degree of lip-voice synchrony, resulting in a stronger correlation than natural videos. In contrast, our model achieves a uniform distribution of audio-visual correlation across all four types of deepfake generation techniques. Note that for clarity in comparison, we have unified the scales of the x and y axes across all figures.}
 \label{motivation}
\end{figure}

\begin{table*}[t]
\centering
\caption{Quantitative comparison of various Deepfake datasets.}

\begin{tabular}{ccccc}
\hline
\multirow{2}{*}{Dataset} & \multirow{2}{*}{Real/Fake} & \multirow{2}{*}{\shortstack{Cross-modal \\ Generation Methods}} & \multirow{2}{*}{\shortstack{Along with \\ Audio}} & \multirow{2}{*}{\shortstack{Public \\ Available}} \\
& & & & \\
\hline 
UADFV~\cite{yang2019exposing} (2018) & $49 / 49$ & 0 & $\times$ & $\times$ \\
Deepfake-TIMIT~\cite{korshunov2018deepfakes} (2018) & $0 / 620$ & 0 & $\checkmark$ & $\checkmark$ \\
FaceForensics++~\cite{rossler2019faceforensics++} (2019) & $1,000 / 4,000$ & 0 & $\times$ & $\checkmark$ \\
Google DFD~\cite{googledfd} (2019) & $363 / 3,068$ & 0 & $\times$ & $\checkmark$ \\
Celeb-DF~\cite{li2020celeb} (2020) & $890 / 5,639$ & 0 & $\times$ & $\checkmark$ \\
DeeperForensics~\cite{jiang2020deeperforensics} (2020) & $10,000 / 50,000$ & 0 & $\times$ & $\checkmark$ \\

DFDC~\cite{dolhansky2019deepfake} (2019) & $23,654 / 104,500$ & 0 & $\checkmark$ & $\checkmark$ \\
FakeAVCeleb~\cite{khalid2021fakeavceleb} (2021) & $500 / 19,500$ & 1 & $\checkmark$ & $\checkmark$ \\
DefakeAVMiT~\cite{yang2023avoid} (2023) & $540 / 6,480$ & 3 & $\checkmark$ & $\times$ \\
\hline 
CMDFD (Ours) & $1,000 / 8,000 $ & 4 & $\checkmark$ & $\checkmark$ \\
\hline
\end{tabular}
\label{classfiy_compare}
\end{table*}

While existing deepfake detection methods have shown promising generalizability within specific modalities~\cite{yu2022focus,haliassos2021lips,zheng2021exploring}, their capability of handling deepfakes generated by various cross-modal forgery methods is still under-explored. Furthermore, most existing cross-modal detection methods~\cite{zhou2021joint,chugh2020not,feng2023self} identify fake videos by capturing out-of-sync or audiovisual mismatches. The fundamental assumption behind this is that deepfake videos always exhibit more audio-visual inconsistencies due to separate modality manipulation compared to natural videos. However, it is not always the case that deepfakes exhibit more obvious audio-visual misalignments compared to real videos. As shown in the top row of Fig.\ref{motivation}, we train a deepfake detection baseline based on a synchronization model~\cite{tao2021someone} and visualize its audio-visual correlation across various types of cross-modal deepfakes. We can see that deepfakes created by talking head generation methods~\cite{zhou2020makelttalk,wang2021audio2head} exhibit lower audio-visual similarity than real videos. However, deepfakes produced by lip-sync
generation methods~\cite{prajwal2020lip,cheng2022videoretalking} often display stronger audio-visual correlations than real videos. This observation suggests that relying on audio-visual synchronization alone cannot provide reliable and generalizable forgery patterns to distinguish various types of cross-modal deepfakes. 

In this work, we propose a deepfake detector capable of generalizing across a wider range of cross-modal deepfakes. We introduce a correlation distillation task to disentangle audio-visual mismatch biases from intrinsic forgery patterns instead of solely relying on measuring the synchronization between audio-visual pairs. In this task, audio speech recognition (ASR) and visual speech recognition (VSR) models are utilized as teacher models, providing soft labels for audio-visual correlation in the level of speech content. We contend that speech recognition models can offer an inherent, content-level cross-modal correlation for videos, regardless of whether they are real or fake. ASR and VSR independently produce probability distributions over the spoken characters. This process maps frame-wise features from different modalities onto the same character set, thereby enabling us to measure a temporal-level cross-modal correlation that can serve as soft labels. These labels are a proxy for assessing the synchronization quality between the audio and visual modalities. By doing so, we guide the framework to model fine-grained synchronization correlations from a content-based perspective. Furthermore, we observe that current multimodal deepfake datasets cover few cross-modal forgery methods. This limitation hinders the development of a generalizable cross-modal deepfake detector. To address this issue, we introduce the Cross-Modal Deepfake Dataset (CMDFD), which encompasses multiple forgery methods, extending beyond lip-sync generation to include more talking head generation methods. 

Our main contributions are summarized as follows: \textbf{(1)} We propose a detection method designed to combat a wider variety of cross-modal deepfakes. Our scheme explicitly learns cross-modal correlation from a content perspective, thereby preventing the model from overfitting based on whether the audio-visual elements are synchronized. \textbf{(2)} We introduce a new benchmark dataset, the Cross-Modal Deepfake Dataset (CMDFD), which contains diverse types of cross-modal deepfakes, including lip-sync and talking head generation, to address the lack of diversity in existing experimental deepfake datasets.  
\textbf{(3)} Extensive experiments demonstrate that our method's generalizability surpasses that of state-of-the-art methods. The correlation learned by our framework exhibits a uniform distribution across multiple cross-modal forgeries (as shown in Fig.~\ref{motivation}), further illustrating our ability to capture more reliable forgery clues.

\section{Related Work}
 \begin{figure*}[h]
 \centering
\includegraphics[width=0.9\textwidth{},keepaspectratio]{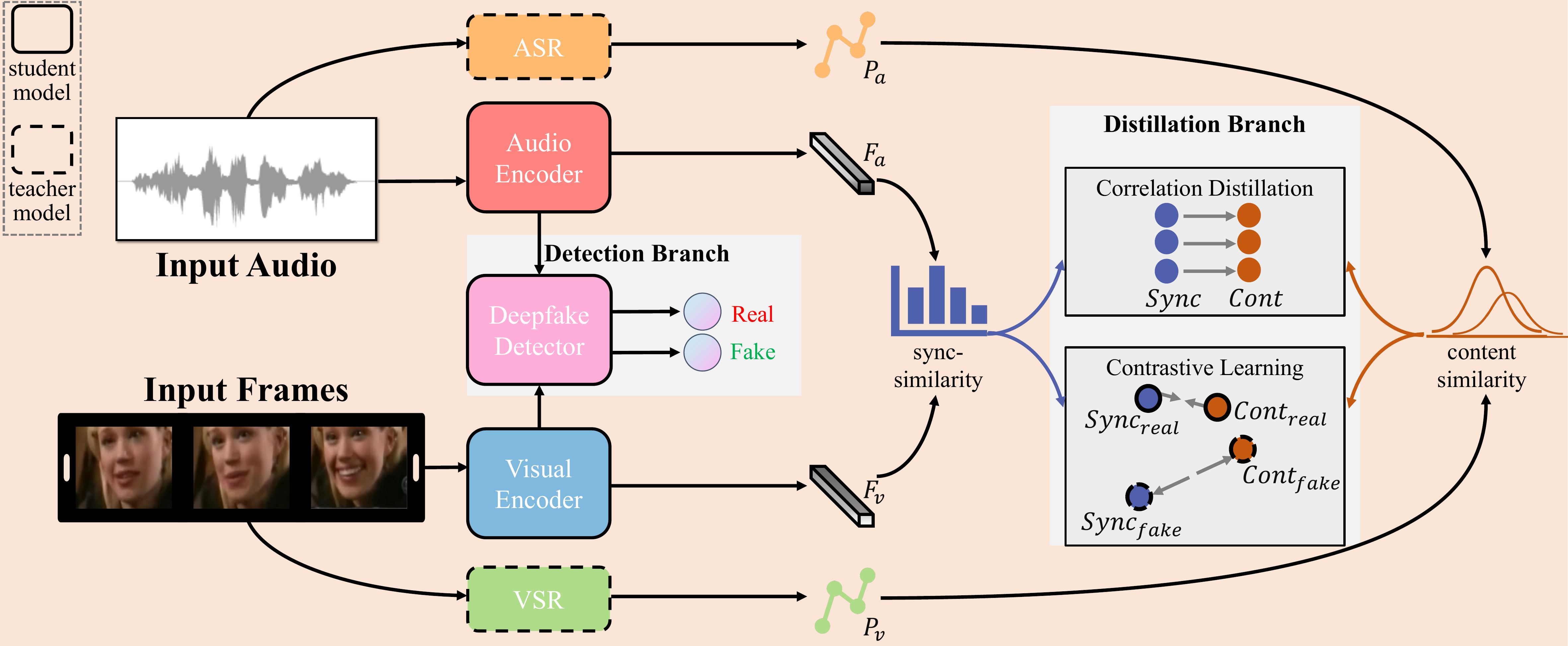}
 \caption{An overview of our method. The overall framework comprises two branches: a detection branch for deepfake prediction and a distillation branch dedicated to cross-modal correlation learning.
 }
 \label{overview}
\end{figure*}

\subsection{Deepfake Generation and Datasets}
The earliest deepfake techniques typically refer to forgery methods focused on manipulating facial information, such as face reenactment~\cite{nirkin2019fsgan,thies2016face2face} and face swap~\cite{korshunova2017fast,perov2020deepfacelab}. Consequently, early deepfake datasets mainly contain deepfakes generated by these methods, exemplified by FaceForensics++~\cite{rossler2019faceforensics++} and DFDC~\cite{dolhansky2019deepfake}. Recently, deepfakes have evolved into multimodal forms, encompassing not only visual deepfakes but also audio deepfakes created by techniques like voice cloning or voice conversion, as well as cross-modal deepfakes. In this context, cross-modal deepfakes refer to those forgeries where the visuals are generated and driven by specific audio inputs. FakeAVCeleb~\cite{khalid2021fakeavceleb} is a widely-used multimodal dataset with such deepfakes. However, it is limited to lip-sync forgeries generated by Wav2Lip~\cite{prajwal2020lip}. Recently, numerous innovative cross-modal forgery methods have emerged. For instance, VideoReTalking~\cite{cheng2022videoretalking} has achieved highly synchronized lip-sync generation. Additionally, talking head generation methods like Audio2Head~\cite{wang2021audio2head} and MakeItTalk~\cite{zhou2020makelttalk} can modify the lip region and head motion. In this work, we aim to encompass these forgery types in our dataset. The comparison of the existing deepfake datasets is presented in Table \ref{classfiy_compare}.

\subsection{Deepfake Detection}
As new types of deepfakes continually emerge, enhancing the detection generalizability has become a focal issue in current research. LipForensics~\cite{haliassos2021lips} focuses on lip region artifacts by utilizing a pre-trained lip-reading model. FTCN~\cite{zheng2021exploring} explores temporal coherence for more general face forgery detection. However, these studies focus their efforts on combating various types of visual deepfakes. To detect multimodal deepfake videos, audio-visual feature fusion is proposed in~\cite{muppalla2023integrating} for fine-grained deepfake detection. AVAD~\cite{feng2023self} detects cross-modal deepfakes by predicting synchronization features pre-learned via an external audio-visual synchronization model. JointAV~\cite{zhou2021joint} exploits synchronization patterns of audiovisual
pairs to detect deepfake videos. Nonetheless, the extent to which detectors can generalize across different types of cross-modal deepfakes remains underexplored in these approaches.
\section{Method}
The overview of our method is depicted in Fig.~\ref{overview}. We develop a multi-task learning framework with two branches to model potential cross-modal correlations during the forgery prediction process explicitly. 
\subsection{Deepfake Detection Branch}
Given a video, we extract the audio into a Mel-frequency cepstral coefficients (MFCCs) vector and detect the face crops in the visual frames. These extracted features are fed into the audio and visual encoders, yielding the audio embedding \( F_a \) and the visual embedding \( F_v \). Then, these embeddings are fed into the detector for deepfake detection. The detector comprises two cross-attention modules as introduced in~\cite{tao2021someone}, where \(F_a\) and \(F_v \) serve as the source sequences for each module, respectively. Subsequently, the outputs of each module are concatenated and fed into a fully connected layer to perform binary detection. This branch is optimized by cross-entropy loss, where \( s_i \) and \( y_i \) denote each video's predicted label and ground truth label, respectively.
\begin{equation}
\mathcal{L}_{c l s}=\mathcal{L}_{c e}\left(y_i, s_i\right)
\end{equation}
\subsection{Correlation Distillation Branch}

Parallel to the detection branch, the second branch of our framework is dedicated to cross-modal correlation distillation. This process involves the explicit extraction of the audio-visual correlation. Upon obtaining the embeddings \( F_a \) and \( F_v \), we calculate their cosine similarity of each timeframe to quantify the degrees of synchronization. To constrain the learning of this branch, we incorporate ASR and VSR as teacher models. The rationale behind this approach is that both ASR and VSR independently produce probability distributions over the spoken characters. This process can map frame-wise features from different modalities into the same character, enabling us to measure a temporal level of content correlation across modalities. 

Therefore, we calculate the Jensen-Shannon (JS) divergence between the predictions of the ASR and VSR models to assign soft pseudo-labels to denote audio-visual correlations. These labels serve as a proxy for the synchronization quality between the audio and visual modalities. By doing so, we guide the framework to model fine-grained synchronization correlations from a content perspective. This branch is optimized by cross-entropy-based distillation loss below, where \(P_a, P_v\) denote the probabilities of ASR and VSR predictions, respectively.
\begin{equation}
\small
\mathcal{L}_{dist}=\mathcal{L}_{c e}\left[\frac{cos\_sim \left(F_a, F_v\right)+1}{2}, 1-JS\left(P_a, P_v\right)\right]
\end{equation}

We design a joint-modal contrastive learning loss to model the uniform potential correlation further. This approach is distinctive as the contrast is not applied directly between two modalities but at a joint-modal level, aiming to constrain the representations of content and synchronization information. Considering that in the generative process of deepfakes, forgery methods could design losses to impose direct constraints on the similarity between visual and audio features, they often fail to maintain alignment with the semantic and temporal information of the entire video. Therefore, the representations of content and synchronization information in natural videos are posited to be inherently closer to each other than those in fake videos. In light of this, we propose the following contrastive loss to narrow the gap between content and synchronization representations in genuine videos within the latent correlation space and to widen it in fake videos as shown in Fig.~\ref{overview}. Given the content correlation measured by JS divergence, denoted by \( Cont \), and the synchronization similarity from our second branch, denoted by \( Sync \), we conduct the contrastive learning as follows, where \(d_i\) denotes the Euclidean distance, and margin is a hyper-parameter:
\begin{equation}
\small
L_{contra}=\left(y_i\right)\left(d_i\right)^2+\left(1-y_i\right) \max \left(\operatorname{margin}-d_i, 0\right)^2
\end{equation}
\begin{equation}
d_i=\|  Cont -  Sync  \|_2
\end{equation}

The overall framework is optimized using the following loss function, where \textit{N} represents the number of training videos, and \textit{T}  denotes the time dimension of video frames. The correlation distillation and contrastive learning are performed at each frame step to extract fine-grained synchronization and content correlation.
\begin{equation}
L= -\frac{1}{N} \sum_{i=1}^N(L_{cls}+\frac{1}{T}\sum_{t=1} L_{dist}+ \frac{1}{T}\sum_{t=1}L_{contra})
\end{equation}
\section{Cross-Modal Deepfake Dataset}

\begin{figure}[t]
 \centering
\includegraphics[width=1.0\columnwidth]{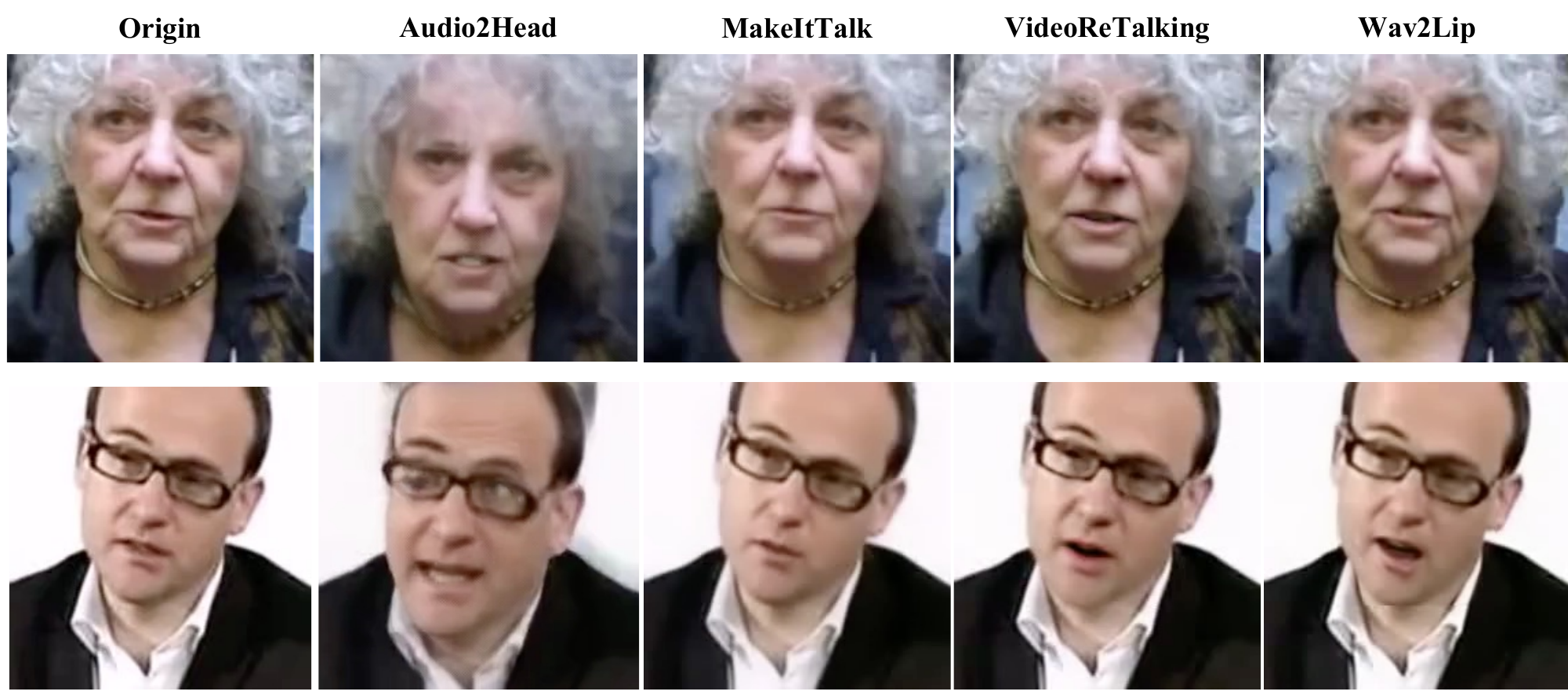}
 \caption{Example video frames in CMDFD.
 }
 \label{dataset}
\end{figure}

Current deepfake datasets are dominated by visual forgery, and limited cross-modal generation methods are incorporated. Our CMDFD dataset aims to include more cross-modal forgery methods to foster the research in generalizable detection, encompassing lip-sync generation and talking head generation forgeries. We gather real videos from the VoxCeleb2~\cite{prajwal2020lip} and manually select 1,000 identities with clear audio speech. Each identity is subjected to manipulation using four forgery methods, including two lip-syncing models and two talking head generation tools: (1) Wav2Lip (W2L)~\cite{prajwal2020lip}, which generates lip-sync videos adaptable to any target speech or language; (2) VideoReTalking (VRT)~\cite{cheng2022videoretalking}, a more recent lip-sync generation method known for enhanced photo-realism; (3) Audio2Head (A2H)~\cite{wang2021audio2head}, which creates talking heads with natural head motion; and (4) MakeItTalk (MIT)~\cite{zhou2020makelttalk}, a technique that manipulates both the lips and the nearby facial region for speaker-aware talking head generation. Examples of these cross-modal deepfakes are illustrated in Fig.~\ref{dataset}. Notably, for each selected visual identity, we manipulated it using another person's speech and the original audio of the identity itself. This replicates a common real-life scenario where a person's voice is employed to manipulate their visual materials. 
Consequently, this results in 2,000 videos for each generation method, yielding a total of 8,000 fake videos. 

\section{Experiment}

\begin{table*}[h!]
\centering
\caption{Generalization results across various forgeries within FakeAVCeleb. The metrics are AUC scores (\%).}
\begin{tabular}{c|c|c|c|c|c|c|c}
\hline 
{Method} & {Modality} 
& RVFA & {FVRA-W2L} & {FVFA-FS} & {FVFA-GAN} & {FVFA-W2L} & AVG-FV \\
\hline
Xception~\cite{rossler2019faceforensics++}  & $\mathcal{V}$  & - & 88.3 & 93.5 & 68.5 & 91.0 & 85.3 \\

LipForensics~\cite{haliassos2021lips}  & $\mathcal{V}$  & - & \textbf{97.7} & 99.9 & 68.1 & 98.7 & 91.1 \\
VQ-GAN~\cite{esser2021taming} & $\mathcal{V}$  & - & 53.8 & 51.2 & 61.1 & 54.4 & 55.1 \\

FTCN~\cite{zheng2021exploring} & $\mathcal{V}$ & -  & 97.4 & \textbf{100.} & 78.3 & 96.5 & 93.1 \\

RealForensics~\cite{haliassos2022leveraging}  & $\mathcal{V}$  & - & 93.0 & 99.1 & \textbf{99.8} & \textbf{96.7} & \textbf{97.1} \\
\hline
AVBYOL~\cite{grill2020bootstrap}  & $\mathcal{A V}$ & 50.0 & 61.3 & 80.8 & 33.8 & 61.0 & 59.2 \\

JointAV~\cite{zhou2021joint}  & $\mathcal{A V}$  & 73.3 & \textbf{97.4} & 99.7 & 55.4 & \textbf{100.} & 88.1 \\
AVAD~\cite{feng2023self} & $\mathcal{A} {\mathcal{V}}$ & 71.6 & 93.7 & 95.8 & 94.3 & 94.1 & 94.5 \\

Ours & $\mathcal{A} {\mathcal{V}}$ & \textbf{76.9} & 95.4 & \textbf{99.9} & \textbf{99.9} & \textbf{100.} & \textbf{98.8}\\
\hline 
\end{tabular}
\label{sota}
\end{table*}

\begin{table*}[h!]
\centering
\caption{Generalization results from FakeAVCeleb to CMDFD. The metrics are AUC scores (\%).}
\label{your-label-here}
\begin{tabular}{l|c|cc|cc|cc|cc|c}
\hline
Method   & FAV   & A2H   & A2H\_S & VRT   & VRT\_S & W2L   & W2L\_S & MIT   & MIT\_S & AVG   \\
\hline
AVAD~\cite{feng2023self}     & 90.34 & 28.72 & 33.22  & 64.93 & 64.00  & 90.60 & 89.62  & 11.68 & 17.10  & 54.93 \\
JointAV~\cite{zhou2021joint}   & 91.97 & 42.71 & 43.00  & 48.78 & 49.18  & 88.30 & 88.30  & 45.08 & 45.12  & 60.27 \\

Ours & \textbf{98.86} &\textbf{68.87} & \textbf{67.97} & \textbf{88.19}& \textbf{87.39} & \textbf{96.81} & \textbf{95.50} & \textbf{89.80} & \textbf{89.61} & \textbf{87.00} \\
\hline
\end{tabular}
\label{repro}
\end{table*}

\begin{table*}[h!]
\caption{Ablation Studies in AUC scores (\%).}
\centering
\begin{tabular}{l|c|cc|cc|cc|cc|c}
\hline
Method & FAV & A2H & A2H\_S & VRT & VRT\_S & W2L & W2L\_S & MIT & MIT\_S & AVG \\
\hline
{w/o Cons\&Dis} & 78.54 & 66.63 & 66.46 & 62.66 & 62.18 & 58.74 & 58.65 & 85.28 & 85.46 & 69.40 \\

w/o Cons & 98.15 & 66.99 & 66.86 & 87.99 & 87.34 & 96.27 & 94.30 & 88.66 & 88.60 & 86.13 \\

Ours & \textbf{98.86} &\textbf{68.87} & \textbf{67.97} & \textbf{88.19}& \textbf{87.39} & \textbf{96.81} & \textbf{95.50} & \textbf{89.80} & \textbf{89.61} & \textbf{87.00} \\
\hline
\end{tabular}
\label{ablation}
\end{table*}

\subsection{Settings}

\textbf{Implementations.} 
The audio encoder in our method is a 2D ResNet34 network introduced in~\cite{chung2020defence}, and the visual encoder contains a visual frontend and the visual temporal network employed ~\cite{tao2021someone}. The initial learning rate is set to $10^{-4}$, and it is decreased by 5\% for every epoch. The audio track is resampled at 16 kHz, and the video is set to 25 fps. Faces are cropped to a size of $112 \times 112$. The feature dimension of \( F_a, F_v \) is set as 128. The margin of $L_{contra}$ is 1.0. We use the publicly available speech recognition model\footnote{\ninept{\url{https://github.com/lordmartian/deep_avsr}}} as the teacher model.

\noindent\textbf{Datasets.} Besides the proposed CMDFD, we utilize the commonly used FakeAVCeleb (FAV) dataset, an audiovisual dataset containing multiple forgery types. In addition to cross-modal forgeries generated by Wav2Lip (W2L), it includes audio deepfakes and visual deepfakes created using FaceSwap (FS)~\cite{korshunova2017fast} and FaceGAN (GAN)~\cite{nirkin2019fsgan}. To demonstrate the detector's generalization performance, we further establish two evaluation protocols: generalization within and across datasets. For the first protocol, we follow the settings in AVAD~\cite{feng2023self} to evaluate detectors on unseen manipulation types within the FAV dataset. This is implemented by conducting five leave-one-out generalization tests based on the generation methods and forgery modalities in FAV, including \textbf{RVFA}, \textbf{FVRA-W2L}, \textbf{FVFA-FS}, \textbf{FVFA-GAN}, and \textbf{FVFA-W2L}. The prefix denotes the forgery modality (for example, RVFA denotes real video fake audio), and the suffixes are abbreviations of the forgery methods. We also report
the average performance over four fake video (FV) categories in AVG-FV. For the second protocol, the detectors are trained on the entire FAV dataset and tested on each forgery type within our CMDFD to examine generalization capabilities across various cross-modal forgery types specifically.

\subsection{Detection Generalization Comparison}
We first compare our method with previous deepfake detectors under the first evaluation protocol. Table~\ref{sota} shows that our method demonstrates the best performance, achieving an average AUC of 98.8\% on all deepfake types with visual forgeries, and surpassing other methods in most forgery types. Additionally, the superior generalization performance of our model on audio-only deepfakes (RVFA) shows that it can detect not only cross-modal deepfakes but also generalize well to unimodal forgeries. Even when a single modality is forged, suspicious cross-modal artifacts are left behind. This underscores the importance of modeling intrinsic cross-modal patterns for detecting various types of forgeries.

Next, we reproduce two open-source baselines dedicated to capturing cross-modal clues and compare them with our method under the second evaluation protocol. The results are presented in Table~\ref{repro}, where entries marked with the ``\_S'' suffix indicate generation types that use the subject's own voice. All methods exhibit excellent performance on W2L forgeries, which are the seen forgery type in the FAV training set. However, the overall results indicate the challenge of generalizable detection across various cross-modal deepfakes. A detailed examination of the classification outcomes reveals that both AVAD and JointAV tend to classify unseen forgery types as genuine. This is because both methods rely on audio-visual synchronization to discern deepfake videos. In contrast, our method models uniform cross-modal correlations and effectively avoids overfitting to audio-visual synchronization patterns. This approach enables our model to perform superior generalization across all types of forgeries consistently. Further, almost all methods achieve slightly lower performance on forgeries generated with the subject's own voice compared to those generated with different voices. 
This indicates that deepfake videos with more congruent audio-visual components are more challenging to detect.

\subsection{Ablation Study}
To evaluate the contribution of each component within our proposed framework, we present an ablation study in Table~\ref{ablation}. We denote the baseline configuration as \textbf{w/o Cons\&Dis}, which refers to a simplified model that only includes the detection branch based on the encoders and detector. The second variant, \textbf{w/o Cons}, represents our multi-task framework comprising both the forgery detection and correlation distillation branches without using contrastive learning in the latent correlation space. The ablation results highlight the significance of each component in our framework. The baseline (w/o Cons\&Dist) achieves lower performance across all datasets, with an average AUC of 69.40\%, highlighting the distillation branch’s necessity for improved performance. Including the correlation distillation branch (w/o Cons) leads to substantial improvements, underpinning the efficacy of capturing intrinsic correlations between audio and visual modalities for deepfake detection. Integrating all components, our method achieves the best performance with an average score of 87.00\%. Overall, the experiment affirms the advantage of our method in enhancing the model's ability to distinguish between genuine and multiple cross-modal deepfakes.

\section{Conclusion}
Our work advances the generalizable detection of deepfakes by integrating explicit learning of cross-modal correlations. By incorporating a novel audio-visual correlation distillation branch into the detection pipeline, our approach effectively captures fine-grained synchronization correlations at the content level. The experiment results highlight the significance of modeling uniform cross-modal correlation across various forgeries for generalizability enhancement. We also propose a CMDFD benchmark dataset with diverse cross-modal deepfakes to promote the comprehensive evaluation of multi-modal deepfake detectors. Our future work will focus on combining complementary audio-visual clues in detection, aiming to achieve a more balanced performance across various forms of forgery.


\small{
\bibliographystyle{IEEEtranS}
\bibliography{icme2023template}
}
\end{document}